\documentclass[10pt,twocolumn,letterpaper]{article}

\usepackage{cvpr}
\usepackage[table]{xcolor}
\usepackage{array}
\usepackage{multirow}
\usepackage{algorithm}
\usepackage{algorithmic}
\usepackage{amsmath}
\usepackage{amssymb}
\usepackage{pgfplots}
\pgfplotsset{compat=1.18}
\definecolor{cvprblue}{rgb}{0.21,0.49,0.74}
\usepackage{graphicx}
\usepackage{array}
\usepackage{tcolorbox}
\usepackage{xcolor}
\usepackage{booktabs}
\usepackage[pagebackref,breaklinks,colorlinks,allcolors=cvprblue]{hyperref}

\def\paperID{5}
\def\confName{CVPR}
\def\confYear{2026}

\title{Towards Responsible Multimodal Medical Reasoning via Context-Aligned Vision-Language Models}

\author{
\textbf{Sumra Khan}$^1$ \quad 
\textbf{Sagar Chhabriya}$^2$ \quad 
\textbf{Aizan Zafar}$^3$ \quad 
\textbf{Sheeraz Arif}$^1$ \\
\textbf{Amgad Muneer}$^4$ \quad 
\textbf{Anas Zafar}$^4$ \quad 
\textbf{Shaina Raza}$^5$ \quad 
\textbf{Rizwan Qureshi}$^{1*}$\\[8pt]
\textsuperscript{1}Department of Computer Science, Salim Habib University, Karachi, Pakistan \\
\textsuperscript{2}Computer Science, Institute of Business Administration Sukkur, Pakistan \\
\textsuperscript{3}Center for Research in Computer Vision, University of Central Florida, USA \\
\textsuperscript{4}The University of Texas MD Anderson Cancer Center, USA \\
\textsuperscript{5}Toronto Metropolitan University, Vector Institute, Canada \\
}

\begin{document}
\maketitle

\vspace*{-0.8cm}
\begin{abstract}
Medical vision–language models (VLMs) show strong performance on radiology tasks but often produce fluent yet weakly grounded conclusions due to over-reliance on a dominant modality. We introduce a context-aligned reasoning framework that enforces agreement across heterogeneous clinical evidence before generating diagnostic conclusions. The proposed approach augments a frozen VLM with structured contextual signals derived from radiomic statistics, explainability activations, and vocabulary-grounded semantic cues. Instead of producing free-form responses, the model generates structured outputs containing supporting evidence, uncertainty estimates, limitations, and safety notes. We observe that auxiliary signals alone provide limited benefit; performance gains emerge only when these signals are integrated through contextual verification. Experiments on chest X-ray datasets demonstrate that context alignment improves discriminative performance (AUC 0.918→0.925) while maintaining calibrated uncertainty. The framework also substantially reduces hallucinated keywords (1.14→0.25) and produces more concise reasoning explanations (19.4→15.3 words) without increasing model confidence (0.70→0.68). Cross-dataset evaluation on CheXpert further reveals that modality informativeness significantly influences reasoning behavior. These results suggest that enforcing multi-evidence agreement improves both reliability and trustworthiness in medical multimodal reasoning, while preserving the underlying model architecture.
\end{abstract}
\section{Introduction}
\label{sec:intro}
Vision-language models (VLMs) are increasingly explored for clinical decision support, including medical visual question answering, report generation, and diagnostic assistance \cite{hartsock2024vision, ye2025multimodal}. Despite strong performance, these systems often produce fluent yet unjustified conclusions that are not grounded in visual evidence or contradict medical knowledge \cite{aljohani2025comprehensive, huang2025survey}. This phenomenon, commonly referred to as \emph{medical hallucination}, remains a major barrier to safe deployment \cite{zhu2025trust}, \cite{kalpelbe2025vlm}. Existing evaluation protocols primarily measure answer accuracy, which fails to detect reasoning inconsistencies such as incorrect anatomical attribution or incompatible clinical relations \cite{zafar2026beyond}.

Clinical diagnosis, however, is not a single-modality prediction problem but a multi-evidence reasoning process \cite{li2025aor, nensa2025future,qureshi2025thinking}. Radiologists integrate heterogeneous sources of information: image appearance, spatial localization, statistical patterns, terminology consistency, and uncertainty awareness. When these sources disagree, clinicians defer conclusions rather than produce confident answers \cite{ardic2025emerging, dass2025multimodal}. Current VLMs lack this behavior because predictions are generated without requiring agreement across independent evidence sources \cite{llavamed,biomedgpt}.

We hypothesize that unreliable medical reasoning arises from missing \emph{context verification} rather than insufficient model capacity. To address this, we introduce context-aligned medical reasoning, a framework that enforces agreement across heterogeneous clinical evidence before generating conclusions. The model is augmented with structured contextual signals derived from radiomic statistics, explainability activations, and vocabulary-grounded semantic cues, and produces structured outputs containing evidence, uncertainty, limitations, and safety statements.

This formulation converts prediction from generative inference into constrained decision making. The model must justify its conclusion through multiple independent evidence channels and explicitly communicate uncertainty, preventing confident outputs unsupported by context. The system therefore behaves conservatively while preserving discriminative performance.

Experiments on chest X-ray datasets demonstrate improved predictive accuracy together with safer reasoning behavior. The proposed approach reduces unsupported diagnostic claims and consistently produces safety-aware outputs, indicating that contextual agreement improves both reliability and performance. The contributions of this work can be summarized in threefold:
\begin{itemize}
\item We introduce context-aligned multimodal reasoning that requires agreement across heterogeneous clinical evidence sources.
\item We demonstrate that reliability can be improved without modifying model architecture or training objectives, but by changing the decision protocol.
\item We show that enforcing uncertainty and safety communication enables responsible medical AI behavior while maintaining predictive performance.
\end{itemize}%-------------------------------------------------------------------------

%-------------------------------------------------------------------------

\section{Related Work}

\paragraph{Medical Vision--Language Models.}
Recent medical VLMs adapt large multimodal architectures for clinical tasks such as visual question answering and report generation \cite{muneer2025classical}. Systems including LLaVA-Med~\cite{llavamed}, BioMedGPT~\cite{biomedgpt}, and RadFM~\cite{radfm} improve answer accuracy by transferring knowledge from pretrained multimodal models to medical data . However, these approaches primarily optimize final prediction quality and do not enforce that conclusions are supported by verifiable clinical evidence \cite{chen2023medical}. As a result, models may produce fluent yet weakly grounded statements, limiting reliability in safety-critical settings.

\paragraph{Reasoning and Hallucination in Medical AI.}
Recent work highlights hallucination and reasoning inconsistency as a central limitation of medical foundation models~\cite{zhu2025trust,kalpelbe2025vlm, chen2023medical}. Approaches such as chain-of-thought prompting~\cite{cot} and reinforcement learning based alignment~\cite{medr1} attempt to improve reasoning quality, but they typically generate free-form explanations that remain difficult to verify \cite{bose2025visual}. In contrast, clinical decision making relies on agreement across multiple independent evidence sources rather than a single textual rationale.

\paragraph{Grounded and Responsible Medical AI.}
Prior efforts improve trustworthiness through visual grounding, interpretability, or structured outputs \cite{ennab2025advancing}. Phrase grounding and spatial alignment methods encourage localization consistency \cite{deng2025med, huy2025seeing}, while safety-oriented frameworks aim to reduce overconfident predictions \cite{zou2025uncertainty}. Our work differs by enforcing \emph{context agreement} across heterogeneous clinical evidence, requiring predictions to remain consistent with multiple contextual signals and to explicitly communicate uncertainty and limitations. This transforms reasoning from descriptive justification into an auditable decision process.
\section{Methodology}
\subsection{Problem Formulation}

Given a medical study consisting of one or more chest radiographs and an associated radiology report, our goal is to perform structured medical reasoning using a tool-augmented multimodal VLM. Formally, each study is defined as:

\begin{equation}
\mathcal{S} = \{ \mathbf{I}_1, \mathbf{I}_2, \mathbf{R} \}
\end{equation}

where $\mathbf{I}_1, \mathbf{I}_2$ denote frontal and lateral radiographs (when available), and $\mathbf{R}$ denotes the corresponding textual report.

We aim to generate a structured reasoning output:

% \begin{equation}
% \mathbf{Y} = \{ y_{\text{imp}}, y_{\text{evid}}, y_{\text{unc}}, y_{\text{lim}}, y_{\text{safety}} \}
% \end{equation}

% \textcolor{red}{where --- stands for ---}
\begin{equation}
\mathbf{Y} = \{ y_{\text{imp}}, y_{\text{evid}}, y_{\text{unc}}, y_{\text{lim}}, y_{\text{safety}} \}
\end{equation}
where $y_{\text{imp}}$ denotes the diagnostic impression, $y_{\text{evid}}$ the supporting clinical evidence, $y_{\text{unc}} \in [0,1]$ the calibrated uncertainty score, $y_{\text{lim}}$ the stated limitations of the assessment, and $y_{\text{safety}}$ the mandatory safety disclaimer required for responsible clinical deployment. Representing impression, supporting evidence, uncertainty score, limitations, and safety notes.

Unlike standard multimodal reasoning, we augment the VLM with structured diagnostic tools derived from image statistics, explainability signals, and vocabulary-grounded semantic cues.

\begin{figure*}[t]
\centering
\includegraphics[width=\textwidth]{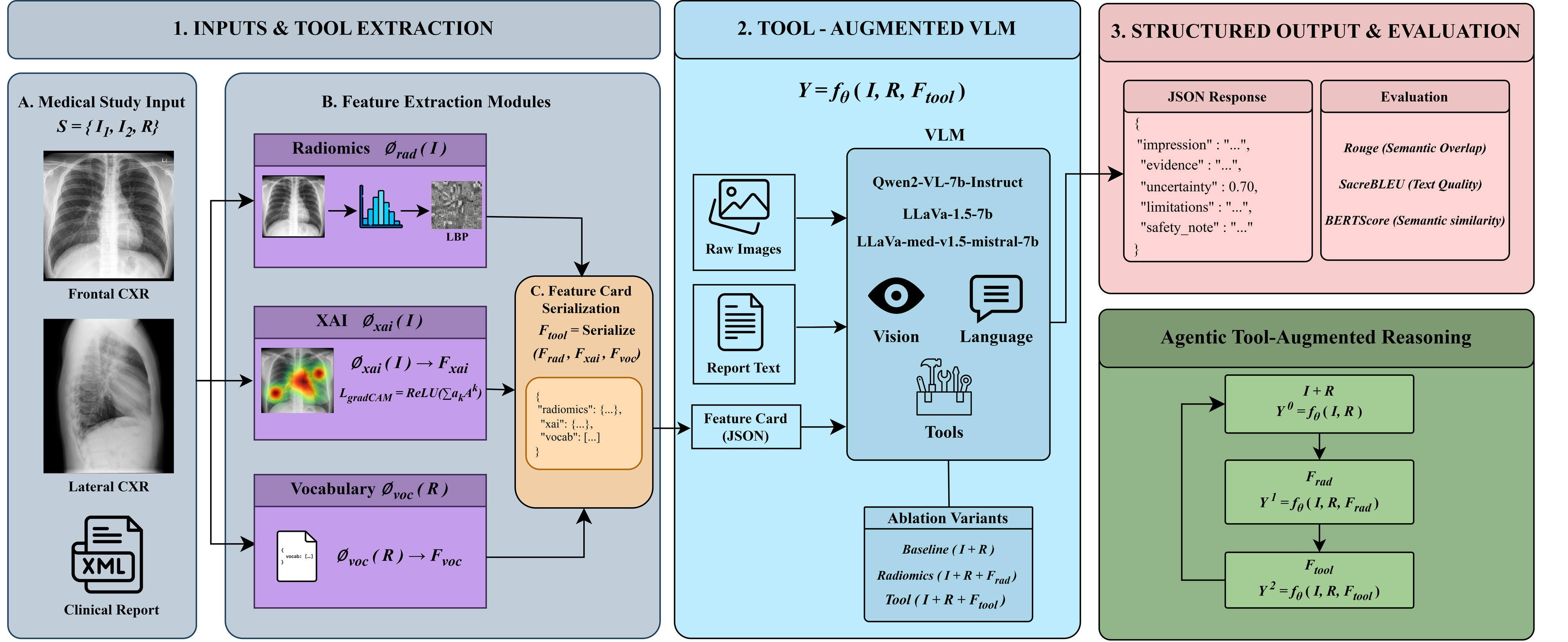}
\caption{
\textbf{Context-aligned reasoning for responsible medical VLMs.}
Conventional medical vision--language models often rely on a dominant modality, which can produce confident but weakly grounded conclusions. 
Our framework augments the VLM with heterogeneous clinical evidence sources, including radiomic statistics, explainability activations, and vocabulary-grounded semantic cues. 
These signals are unified through a \emph{context agreement constraint} before inference, forcing the model to justify predictions across independent evidence channels. 
Instead of generating free-form text, the model produces a structured clinical response containing impression, supporting evidence, calibrated uncertainty, limitations, and safety notes. 
This transforms prediction from generative output into an auditable decision process, improving reliability while preserving performance.
}
\label{fig:framework}
\end{figure*}
\subsection{Three-Stage Context-Aligned Reasoning Framework}

The proposed architecture harmonizes five heterogeneous information streams: raw medical imagery $\mathbf{I}$, unstructured radiology reports $\mathbf{R}$, radiomic inspired descriptors $\mathbf{F}_{rad}$, explainability derived activation statistics $\mathbf{F}_{xai}$, and vocabulary aligned semantic features $\mathbf{F}_{voc}$.

These components are integrated within a \textbf{three-stage context-aligned reasoning framework}: 

\begin{itemize}
\item \textbf{Stage 1: Heterogeneous Context Extraction}
\item \textbf{Stage 2: Context Serialization and Multimodal Fusion}
\item \textbf{Stage 3: Tool-Augmented Agentic Reasoning}
\end{itemize}

The overall reasoning process is modeled as:

\begin{equation}
\mathbf{Y} = f_{\theta}(\mathbf{I}, \mathbf{R}, \mathbf{F}_{rad}, \mathbf{F}_{xai}, \mathbf{F}_{voc})
\end{equation}

where $f_{\theta}$ denotes the pretrained Qwen2-VL model with frozen parameters.

\subsection{Stage 1: Heterogeneous Context Extraction}

\paragraph{Radiomic Intensity and Texture Statistics.}

To provide additional clinical context, we extract low-level radiomic descriptors from the input image $I \in \mathbb{R}^{H \times W}$. First-order intensity statistics are computed using the mean 
$\mu = \frac{1}{HW}\sum_{i,j} I_{ij}$ and variance 
$\sigma^2 = \frac{1}{HW}\sum_{i,j}(I_{ij}-\mu)^2$, together with percentile and intensity range features.

To capture spatial texture patterns, we construct the gray-level co-occurrence matrix $P(i,j,d,\theta)$ and derive standard texture measures including contrast 

$\sum_{i,j}(i-j)^2 P(i,j)$ and homogeneity 

$\sum_{i,j} \frac{P(i,j)}{1+|i-j|}$.

These statistics summarize global intensity distribution and local structural organization, providing complementary evidence to visual embeddings for downstream reasoning.

\subsection{Explainability-Derived Features}

To capture model-driven spatial attention, we compute Grad-CAM maps from a pretrained DenseNet-121 backbone. Given feature activations $A^k$ at the final convolutional layer and class score $y^c$, Grad-CAM weights are:

\begin{equation}
\alpha_k^c = \frac{1}{Z} \sum_{i,j} \frac{\partial y^c}{\partial A_{ij}^k}
\end{equation}

The class activation map is:

\begin{equation}
L_{GradCAM}^c = \text{ReLU} \left( \sum_k \alpha_k^c A^k \right)
\end{equation}

From the normalized activation map, we compute summary statistics:

\begin{equation}
\mathbf{F}_{xai} = \{\text{mean}, \text{max}, \text{entropy}, \text{top-10\% mass}\}
\end{equation}

These quantify spatial concentration and attention entropy.

\subsection{Vocabulary-Grounded Semantic Features}

We extract domain-specific medical terminology from reports using curated radiology vocabulary.

Let $\mathcal{V}$ denote the vocabulary set. For each report:

\begin{equation}
\mathbf{F}_{voc} = \{ v \in \mathcal{V} \mid v \subset \mathbf{R} \}
\end{equation}

These matched concepts serve as structured semantic anchors for reasoning.

\subsection{Stage 2: Context Serialization and Multimodal Fusion}

Structured tool outputs are serialized into a feature card:

\begin{equation}
\mathbf{F}_{tool} = \text{Serialize}(\mathbf{F}_{rad}, \mathbf{F}_{xai}, \mathbf{F}_{voc})
\end{equation}

This unified feature representation synthesizes heterogeneous clinical evidence into a structured multimodal context that can be consumed by the vision-language model.

\subsection{Stage 3: Tool-Augmented Agentic Reasoning}

We use Qwen2-VL-2B-Instruct as our backbone VLM. The model receives multimodal input via a chat template:

\begin{equation}
\mathbf{Y} = f_{\theta}(\mathbf{I}, \mathbf{R}, \mathbf{F}_{tool})
\end{equation}

We evaluate two modes: \textbf{Single-Shot Augmentation} and \textbf{Stepwise Agentic Reasoning}. 

For Single-Shot Augmentation, all tool features are provided simultaneously. For Stepwise Agentic Reasoning, the reasoning proceeds in the following stages:

\begin{align}
\mathbf{Y}^{(0)} &= f_{\theta}(\mathbf{I}, \mathbf{R}) \\
\mathbf{Y}^{(1)} &= f_{\theta}(\mathbf{I}, \mathbf{R}, \mathbf{F}_{rad}) \\
\mathbf{Y}^{(2)} &= f_{\theta}(\mathbf{I}, \mathbf{R}, \mathbf{F}_{tool})
\end{align}

This allows analysis of reasoning refinement and uncertainty shifts.

\subsection{Responsible AI Constraints}

We enforce structured outputs in JSON format and require explicit uncertainty estimation:

\begin{equation}
y_{\text{unc}} \in [0,1]
\end{equation}

The model is instructed to: (i) Avoid definitive diagnosis claims; (ii) Explicitly state limitations and (iii) Provide safety disclaimers. This constraint promotes safer and calibrated reasoning behavior\cite{raza2025responsible}.

Algorithm \ref{alg:overall_framework} formalizes our context aligned reasoning pipeline, detailing the systematic progression from heterogeneous feature extraction to structured decision making. By explicitly defining the tool augmented reasoning phases, the algorithm illustrates how multi evidence integration enforces responsible artificial intelligence constraints prior to generating a final clinical response.

\begin{algorithm}[t]
\caption{Context Aligned Multimodal Reasoning for Medical Vision Language Models}
\label{alg:overall_framework}
\begin{algorithmic}[1]
\REQUIRE Medical study $\mathcal{S} = \{ \mathbf{I}, \mathbf{R} \}$
\REQUIRE Frozen vision language model $f_{\theta}$; Pretrained DenseNet 121
\REQUIRE Medical vocabulary $\mathcal{V}$
\ENSURE Structured clinical response $\mathbf{Y} = \{ y_{\text{imp}}, y_{\text{evid}}, y_{\text{unc}}, y_{\text{lim}}, y_{\text{safety}} \}$

\STATE \(\triangleright\) \textbf{Phase 1: Heterogeneous Context Extraction}
\STATE $\mathbf{F}_{rad} \gets \text{ExtractRadiomics}(\mathbf{I})$ \COMMENT{Intensity $\mu, \sigma^{2}$ and texture $P(i,j)$}
\STATE $L_{GradCAM} \gets \text{ComputeGradCAM}(\mathbf{I})$
\STATE $\mathbf{F}_{xai} \gets \text{DeriveSpatialStatistics}(L_{GradCAM})$ \COMMENT{mean, max, entropy, top mass}
\STATE $\mathbf{F}_{voc} \gets \text{ExtractSemanticAnchors}(\mathbf{R}, \mathcal{V})$

\STATE \(\triangleright\) \textbf{Phase 2: Context Serialization and Alignment}
\STATE $\mathbf{F}_{tool} \gets \text{Serialize}(\mathbf{F}_{rad}, \mathbf{F}_{xai}, \mathbf{F}_{voc})$ \COMMENT{Synthesize unified tool feature card}

\STATE \(\triangleright\) \textbf{Phase 3: Tool Augmented Reasoning}
\IF{reasoning mode is Single Shot}
    \STATE $\mathbf{Y} \gets f_{\theta}(\mathbf{I}, \mathbf{R}, \mathbf{F}_{tool})$ \COMMENT{Immediate multi evidence output}
\ELSIF{reasoning mode is Stepwise Agentic}
    \STATE $\mathbf{Y}^{(0)} \gets f_{\theta}(\mathbf{I}, \mathbf{R})$ \COMMENT{Baseline vision language inference}
    \STATE $\mathbf{Y}^{(1)} \gets f_{\theta}(\mathbf{I}, \mathbf{R}, \mathbf{F}_{rad})$ \COMMENT{Refine reasoning with radiomic context}
    \STATE $\mathbf{Y}^{(2)} \gets f_{\theta}(\mathbf{I}, \mathbf{R}, \mathbf{F}_{tool})$ \COMMENT{Final agreement grounded decision}
    \STATE $\mathbf{Y} \gets \mathbf{Y}^{(2)}$
\ENDIF

\STATE \(\triangleright\) \textbf{Phase 4: Responsible AI Constraint Verification}
\STATE Enforce schema structure and calibrated uncertainty constraint $y_{\text{unc}} \in [0,1]$
\STATE Verify absence of definitive diagnostic claims
\STATE Verify presence of mandatory safety disclaimers
\RETURN $\mathbf{Y}$
\end{algorithmic}
\end{algorithm}

\subsection{Ablation Study}

To quantify modality contribution, we train logistic regression classifiers over feature combinations:

\begin{equation}
\hat{y} = \sigma(\mathbf{w}^\top \mathbf{x})
\end{equation}

where $\mathbf{x}$ corresponds to radiomic, XAI, text embedding, and vocabulary features. We report AUC across modality combinations to assess complementary signal contributions.
\section{Experimental Setup}
Experiments are conducted on a Tesla T4 GPU using half-precision inference. 
Large-scale single-shot experiments are performed on 1,000 studies, while stepwise agentic analysis is conducted on a 50-study subset to examine reasoning dynamics.
\subsection{Datasets}

\textbf{OpenI.} 
We evaluate on the Indiana University Chest X-ray Collection (OpenI), 
which contains paired chest radiographs and detailed free-text radiology reports. 
We construct matched image-report pairs and use 1,000 studies for ablation experiments 
and 50 studies for agentic reasoning analysis. \textbf{CheXpert.}
For cross-dataset validation, we evaluate on CheXpert, 
a large-scale chest X-ray dataset with image-level labels derived from automated report parsing. 
Compared to OpenI, CheXpert contains shorter and more label-centric textual content.

\subsection{Feature Extraction}

Radiomic-inspired features include first-order intensity statistics 
(mean, standard deviation, percentiles) and texture descriptors 
derived from Gray-Level Co-occurrence Matrices (GLCM) and 
Local Binary Patterns (LBP). Explainability signals are extracted using Grad-CAM from a 
pretrained DenseNet-121 backbone. 
We summarize activation maps using mean activation, entropy, 
and spatial concentration metrics. Textual representations are computed using pretrained sentence embeddings.

\subsection{Vision-Language Model}

We use Qwen2-VL-2B-Instruct as a frozen vision-language backbone. 
Structured tool outputs are injected via controlled prompting. 
All outputs are constrained to strict JSON format with numeric uncertainty 
scores between 0 and 1.

\subsection{Evaluation Protocol}

To evaluate multimodal contributions, we utilize logistic regression classifiers trained on various feature combinations and benchmarked by the Area Under the ROC Curve (AUC). Simultaneously, the agentic analysis quantifies uncertainty calibration through mean and variance, measures evidence length by word count, and identifies the presence of safety notes or limitations. We conduct all experiments using a fixed random seed to ensure rigorous reproducibility and consistent results across trials.

\section{Results}
\subsection{Multimodal Abllation Study}

To quantify the contribution of different modalities, we analyze feature combinations using a logistic regression classifier trained on 1,000 studies from the OpenI dataset. The evaluated features include textual embeddings derived from radiology reports, radiomic-inspired image statistics, and explainability-derived (XAI) activation summaries. Performance is measured using the Area Under the ROC Curve (AUC).

\begin{figure}[t]
\centering
\includegraphics[width=\linewidth]{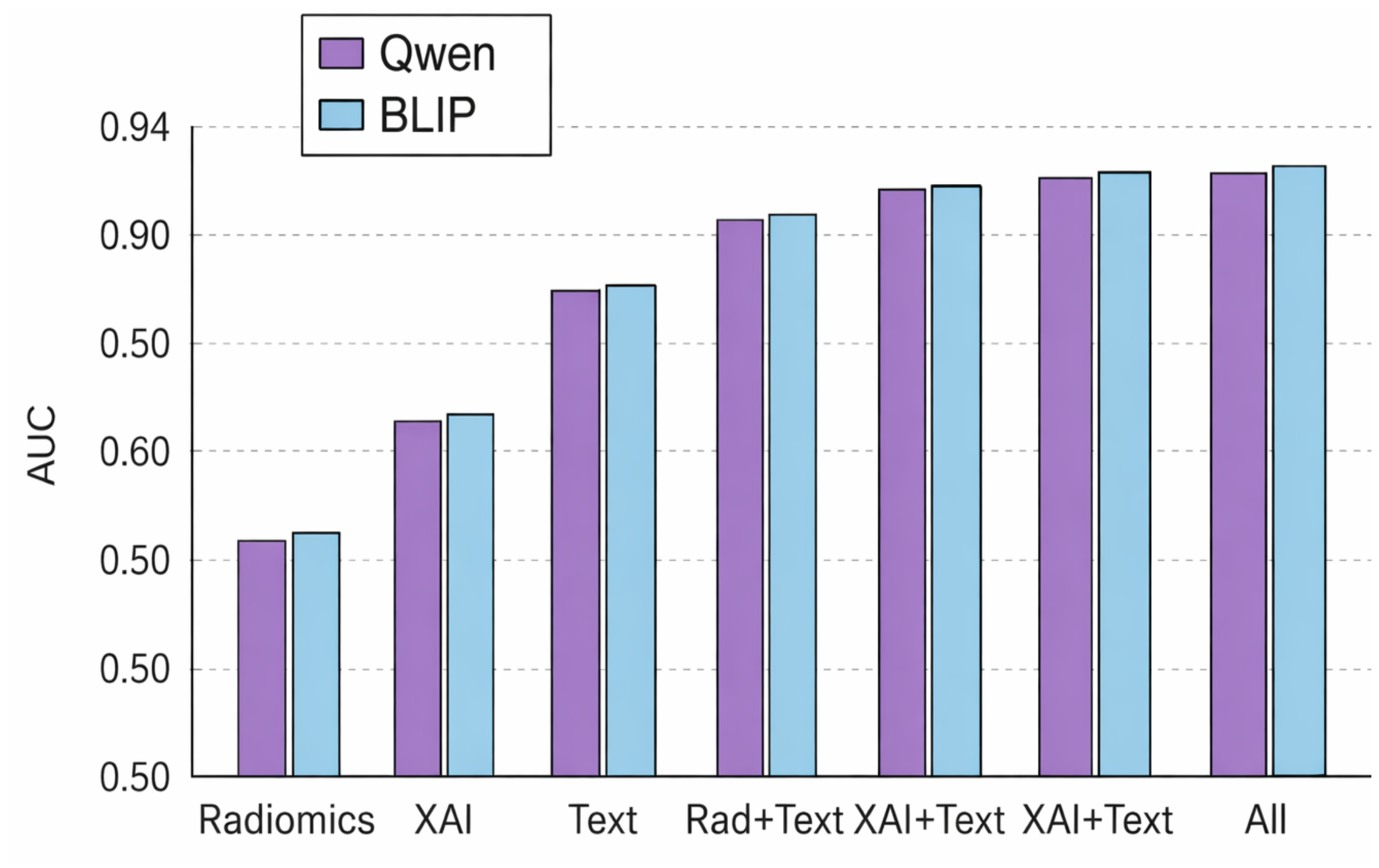}
\caption{
AUC comparison across modality combinations for Qwen and BLIP embeddings on the OpenI dataset. 
Multimodal fusion consistently improves performance compared to single-modality features.
}
\label{fig:ablation_chart}
\vspace{-0.4cm}
\end{figure}

Table~\ref{tab:multimodal_ablation} summarizes the results, while Figure~\ref{fig:ablation_chart} provides a visual comparison of performance across modality combinations. Textual representations provide the strongest predictive signal across both models. Using Qwen embeddings, the text-only configuration achieves an AUC of 0.918, significantly outperforming radiomics-only (0.555) and XAI-only (0.519). A similar trend is observed for BLIP embeddings, where text-only features achieve 0.925 AUC compared to 0.564 for radiomics and 0.514 for XAI.
\begin{table}[t]
\centering
\caption{Multimodal ablation study on the OpenI dataset (AUC).}
\label{tab:multimodal_ablation}
\begin{tabular}{lcc}
\toprule
Feature Configuration & Qwen & BLIP \\
\midrule
Radiomics only & 0.555 & 0.564 \\
XAI only & 0.519 & 0.514 \\
Text only & 0.918 & 0.925 \\
Radiomics + Text & 0.921 & 0.926 \\
XAI + Text & 0.918 & 0.927 \\
Radiomics + XAI + Text & \textbf{0.925} & \textbf{0.930} \\
\bottomrule
\end{tabular}
\end{table}

Despite their weaker standalone performance, image-derived features provide complementary benefits when combined with textual embeddings. For Qwen, incorporating radiomic features improves performance from 0.918 to 0.921, while the full multimodal configuration (radiomics + XAI + text) achieves the best result of 0.925 AUC. A comparable improvement is observed with BLIP embeddings, where the multimodal combination reaches 0.930 AUC.
\begin{table}[ht]
\centering
\small % Keeps font consistent with other academic tables
\caption{Image-centric ablation study evaluating hallucination metrics. \textnormal{HR denotes the Hallucination Rate (average hallucinated keywords per study).}}
\label{tab:image_ablation}
\begin{tabular}{lc}
\toprule
\textbf{Configuration} & \textbf{Hallucination Rate (HR)} $\downarrow$ \\
\midrule
Image only                               & 1.14 \\
Image + Radiomics                        & 1.11 \\
Image + XAI                              & 1.01 \\
Image + Radiomics + XAI                  & 0.60 \\
Image + Text                             & \textbf{0.25} \\
Image + Text + Radiomics + XAI           & 0.28 \\
\bottomrule
\end{tabular}
\end{table}
\begin{table*}[t]
\centering
\caption{Quantitative analysis of agentic reasoning performance across fifty clinical studies. Values represent mean and standard deviation where applicable.}
\label{tab:agentic}
\setlength{\tabcolsep}{12pt} % Adjust horizontal spacing for a cleaner look
\renewcommand{\arraystretch}{1.2} % Better vertical breathing room
\begin{tabular}{lccc}
\toprule
\rowcolor{gray!10} \textbf{Evaluation Metric} & \textbf{Base Conf (Step 0)} & \textbf{Tool Augmented (Step 1)} & \textbf{Absolute Change ($\Delta$)} \\
\midrule
Uncertainty Calibration (mean $\pm$ std) & $0.70 \pm 0.00$ & $0.68 \pm 0.14$ & $\uparrow 0.014$ \\
Evidence Length (total words) & $19.4 \pm 14.0$ & $15.3 \pm 18.0$ & $\downarrow 4.1$ \\
Presence of Limitations & $1.00$ & $0.80$ & $\downarrow 0.20$ \\
Presence of Safety Notes & $1.00$ & $0.80$ & $\downarrow 0.20$ \\
\bottomrule
\end{tabular}
\end{table*}
\begin{table*}[t]
\centering
\caption{\textbf{Responsible-AI evaluation and text quality metrics} on the test split ($N=217$ per variant). 
We report ROUGE (R-1/R-2/R-L) and BERTScore-F1 for generation quality, along with heuristic Responsible-AI indicators: 
PHI leakage rate, unsafe content rate, and uncertainty marker rate. Best values per model (higher is better for ROUGE/BERT; lower is better for PHI/Unsafe) are highlighted in bold.}
\label{tab:rai_results}
\small
\setlength{\tabcolsep}{6pt}
\begin{tabular}{llrccccccc}
\toprule
\textbf{Model} & \textbf{Variant} & \textbf{N} &
\textbf{R-1} $\uparrow$ & \textbf{R-2} $\uparrow$ & \textbf{R-L} $\uparrow$ &
\textbf{BERT} $\uparrow$ &
\textbf{PHI \%} $\downarrow$ & \textbf{Unsafe \%} $\downarrow$ &
\textbf{Unc. rate} \\
\midrule

\multirow{6}{*}{LLaVA-1.5-7B}
& A0\_image\_only           & 217 & 0.0808 & 0.0107 & 0.0494 & 0.7539 & 0.00 & 0.0 & 0.1344 \\
& A1\_radiomics             & 217 & 0.0829 & 0.0079 & 0.0463 & 0.7522 & 0.00 & 0.0 & 0.0661 \\
& A2\_xai                   & 217 & 0.0832 & 0.0100 & 0.0490 & 0.7527 & 0.00 & 0.0 & 0.0929 \\
& A3\_rad\_xai              & 217 & 0.0773 & 0.0062 & 0.0429 & 0.7498 & 0.00 & 0.0 & 0.0269 \\
& A4\_image\_text           & 217 & \textbf{0.4910} & \textbf{0.3190} & \textbf{0.4711} & 0.7640 & 0.92 & 0.0 & 0.0453 \\
& A5\_image\_text\_rad\_xai & 217 & 0.4384 & 0.2825 & 0.4168 & \textbf{0.7646} & 0.92 & 0.0 & 0.0553 \\
\midrule

\multirow{6}{*}{Qwen2-VL-7B}
& A0\_image\_only           & 217 & 0.0951 & 0.0116 & 0.0530 & 0.7778 & 0.00 & 0.0 & 0.1682 \\
& A1\_radiomics             & 217 & 0.1009 & 0.0111 & 0.0514 & 0.7754 & 0.00 & 0.0 & 0.1720 \\
& A2\_xai                   & 217 & 0.0964 & 0.0108 & 0.0509 & \textbf{0.7785} & 0.00 & 0.0 & 0.1375 \\
& A3\_rad\_xai              & 217 & 0.1018 & 0.0118 & 0.0535 & 0.7734 & 0.00 & 0.0 & 0.1352 \\
& A4\_image\_text           & 217 & \textbf{0.4640} & \textbf{0.3023} & \textbf{0.4468} & 0.7791 & 0.92 & 0.0 & 0.0453 \\
& A5\_image\_text\_rad\_xai & 217 & 0.4253 & 0.2753 & 0.4059 & 0.7599 & 0.92 & 0.0 & 0.0399 \\
\bottomrule
\end{tabular}
\end{table*}
\subsection{Image-Centric Reasoning Analysis}

We further evaluate an image-centric setting where visual features serve as the primary modality and additional signals are incrementally integrated, including radiomic statistics, explainability-derived (XAI) cues, and textual report embeddings. As shown in Table~\ref{tab:image_ablation}, the image-only configuration produces the highest hallucination frequency (1.14 on average), indicating that visual features alone are insufficient for reliable clinical reasoning. Incorporating radiomic or XAI features slightly reduces hallucinated outputs (1.11 and 1.01 respectively), while combining radiomics and XAI leads to a larger reduction (0.60).

The most substantial improvement occurs when textual information is introduced. The image+text configuration reduces hallucinated keywords to 0.25 on average, demonstrating the strong semantic grounding provided by radiology reports.
Although, We observe that providing additional features alone does not improve model performance. However, when these features are integrated with contextual information, the model shows consistent performance improvements.
\begin{table}[htbp] % Use 't' for top of page, 'h' for 'here'
    \centering
    \small
    \setlength{\tabcolsep}{4pt}
    \renewcommand{\arraystretch}{1.2}
    
    % The \caption command automates the "Table X:" numbering
\caption{Qualitative visual reasoning example of Qwen.
\textnormal{\textcolor{green!60!black}{Green} text indicates visual grounding;
\textcolor{blue!80!black}{blue} text indicates the structured reasoning chain.}}
    \label{tab:visual_reasoning_example}
    \vspace{6pt}

    \begin{tabular}{|p{0.25\columnwidth}|p{0.7\columnwidth}|}
    \hline
    
    \multicolumn{2}{|p{0.95\columnwidth}|}{
    \textbf{Question:} Based on the chest X-ray, is there evidence of active cardiopulmonary abnormality?
    } \\ \hline
    
    \multicolumn{2}{|p{0.95\columnwidth}|}{
    \textbf{Ground Truth (Radiology Report):} No active cardiopulmonary disease.
    } \\ \hline
    
    \textbf{Input Image} &
    \textbf{Agentic Reasoning and Structured Output} \\ \hline
    
    % Ensure graphicx package is loaded in preamble
    \begin{minipage}{0.23\columnwidth}
        \vspace{4pt}
        \includegraphics[width=\linewidth]{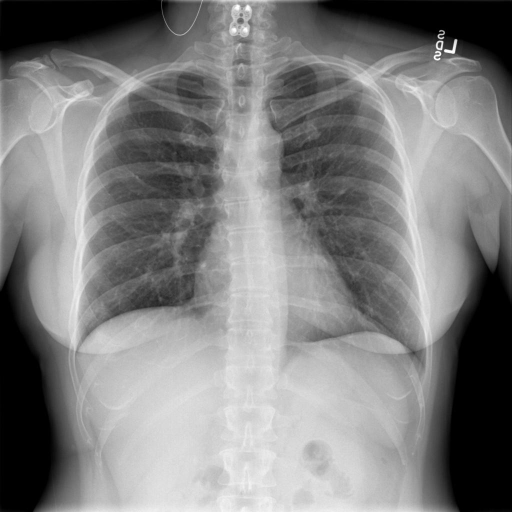}
        \vspace{4pt}
    \end{minipage}
    &
    \begin{minipage}[t]{0.7\columnwidth}
        \vspace{4pt}
        \textbf{Step 1 (Visual Observation):} 
        \textcolor{green}{Clear bilateral lung fields and normal cardiac silhouette.} \\[4pt]
        
        \textbf{Step 2 (Feature Assessment):}
        \textcolor{green}{No focal consolidation, pleural effusion, or pneumothorax identified.} \\[4pt]
        
        \textbf{Step 3 (Structured Reasoning):}
        \textcolor{blue}{Absence of acute radiographic findings suggests no active cardiopulmonary abnormality.} \\[4pt]
        
        \textbf{Step 4 (Uncertainty Calibration):}
        \textcolor{blue}{Numeric uncertainty reflects limited clinical context and single-view imaging.} \\[6pt]
        
        \textbf{Final Structured Output:} \\
        \textbf{Impression:} No obvious radiographic evidence of active cardiopulmonary abnormality. \\
        \textbf{Evidence:} Lung fields clear. Cardiac silhouette within normal limits. \\
        \textbf{Uncertainty\ 0 to 1:} 0.35 \\
        \textbf{Limitations:} Single frontal radiograph without prior comparison. \\
        \textbf{Safety Note:} For research use only; not a substitute for expert interpretation. 
        \vspace{4pt}
    \end{minipage} \\ \hline
    
    \end{tabular}
\end{table}

\begin{table}[htbp]
    \centering
    \small
    \setlength{\tabcolsep}{4pt}
    \renewcommand{\arraystretch}{1.2}
    
    % Automated numbering via \caption. 
    % Note: I've moved the formatting description inside the caption for professional styling.

    \caption{Qualitative visual reasoning example of BLIP.
\textnormal{\textcolor{green!60!black}{Green} text indicates visual grounding;
\textcolor{blue!80!black}{blue} text indicates the structured reasoning chain.}}
    \label{tab:visual_reasoning_example_blip}
    \vspace{6pt}

    \begin{tabular}{|p{0.22\columnwidth}|p{0.73\columnwidth}|}
    \hline
    
    \multicolumn{2}{|p{0.95\columnwidth}|}{
    \textbf{Question:} Based on the chest X-ray and patient history, is there evidence of active cardiopulmonary abnormality?
    } \\ \hline
    
    \multicolumn{2}{|p{0.95\columnwidth}|}{
    \textbf{Ground Truth (Radiology Report):} No active disease.
    } \\ \hline
    
    \textbf{Input Image} &
    \textbf{Agentic Reasoning and Structured Output of BLIP} \\ \hline
    
    % Using a minipage to vertically center the image within the row
    \begin{minipage}{0.22\columnwidth}
        \vspace{2pt}
        \includegraphics[width=\linewidth]{sec/CXR3_IM-1384-1001.png}
        \vspace{2pt}
    \end{minipage}
    &
    \begin{minipage}[t]{0.7\columnwidth}
        \vspace{2pt}
        \textbf{Step 1 (Visual Observation):} 
        \textcolor{green}{Chest radiograph demonstrates clear bilateral lung fields and a normal cardiac silhouette.} \\[2pt]
        
        \textbf{Step 2 (Feature Assessment):}
        \textcolor{green}{No focal consolidation, pleural effusion, or pneumothorax is identified via radiomic feature mapping. The heart and mediastinum appear normal.} \\[2pt]
        
        \textbf{Step 3 (Structured Reasoning):}
        \textcolor{blue}{The absence of acute radiographic findings suggests no active cardiopulmonary abnormality.} \\[2pt]
        
        \textbf{Step 4 (Uncertainty Calibration):}
        \textcolor{blue}{Uncertainty is assigned at a moderate-to-high level due to the patient’s history of renal cell carcinoma, requiring cautious clinical interpretation despite clear imaging findings.} \\[2pt]
        
        \textbf{Final Structured Output:} \\
        \textbf{Impression:} No obvious radiographic evidence of active cardiopulmonary abnormality (Uncertain). \\
        \textbf{Evidence:} The chest X-ray shows clear and expanded lung fields with no active disease. The heart and mediastinum appear normal, with no signs of pulmonary pathology. \\
        \textbf{Uncertainty\ 0 to 1:} 0.7 \\
        \textbf{Limitations:} Assessment is based on a two-view chest radiograph, and radiomics analysis reflects only a single temporal observation. \\
        \textbf{Safety Note:} The patient has a history of renal cell carcinoma; this output should be interpreted by a qualified radiologist.
        \vspace{6pt}
    \end{minipage} \\ \hline
    
    \end{tabular}
\end{table}
\subsection{Agentic Tool-Augmented Reasoning}
To analyze reasoning refinement, we evaluate the agentic reasoning process on 50 randomly sampled studies. Step 0 corresponds to image and report reasoning, while Step 2 includes radiomics, explainability statistics, and vocabulary features. Table~\ref{tab:agentic} summarizes the quantitative analysis. Uncertainty remains globally stable (0.70 vs.\ 0.68), with only a small mean shift ($\Delta = 0.014 \pm 0.052$). Importantly, tool augmentation does not inflate confidence. Evidence length decreases from 19.4 to 15.3 words, indicating more concise reasoning when structured tool signals are available.
\subsection{Cross-Dataset Evaluation}

To evaluate generalization, we apply the same framework to the CheXpert dataset. Radiomic features achieve stronger standalone performance (AUC = 0.71), while text embeddings provide limited predictive signal (AUC = 0.50), reflecting the label-centric and shorter report structure of CheXpert. Unlike OpenI, tool augmentation does not significantly alter uncertainty calibration, and outputs frequently default to a constant uncertainty value of 0.70. These observations indicate that dataset characteristics strongly influence multimodal reasoning dynamics.

\subsection{Qualitative Analysis}

To further illustrate the behavior of the proposed framework, we present representative qualitative examples of agentic reasoning for the Qwen and BLIP variants in Table~\ref{tab:visual_reasoning_example} and Table~\ref{tab:visual_reasoning_example_blip}, respectively. These examples demonstrate the full reasoning pipeline from visual input to structured clinical output.

As shown in these cases, the models first perform visual grounding, identifying clear bilateral lung fields and a normal cardiac silhouette. The framework then proceeds through structured reasoning steps that connect visual findings to a cautious clinical interpretation. The final response adheres to the predefined structured format containing an impression, supporting evidence, calibrated uncertainty, limitations, and a necessary safety note..

\subsection{Responsible AI Evaluation}
Our Responsible AI evaluation results are presented in Table~\ref{tab:rai_results}. We evaluate both generation quality and safety-related indicators across two multimodal models (LLaVA-1.5-7B and Qwen2-VL-7B) under six ablation settings (A0–A5). Text-conditioned variants (A4 and A5) substantially improve report generation quality compared to image-only or feature-only configurations, achieving the highest ROUGE and BERTScore values. In particular, LLaVA-1.5-7B with image–text conditioning (A4) achieves the best lexical overlap with ground truth (ROUGE-L = 0.471), while Qwen2-VL-7B obtains the highest semantic similarity (BERTScore-F1 = 0.779).

From a Responsible AI perspective, all models produce zero unsafe outputs across the evaluation set. PHI detection remains near zero for most variants, with a small number of heuristic triggers in text-conditioned settings, likely due to numerical measurements rather than actual identifiers. Additionally, uncertainty marker rates decrease for text-conditioned variants, suggesting more confident and structured report generation. These results indicate that incorporating textual clinical context significantly improves both report quality and responsible generation behavior.
Table~\ref{tab:rai_results} summarizes generation quality metrics and Responsible AI indicators across all model variants.

\section{Conclusion}
In this work, we introduced a context aligned reasoning framework to address the pervasive challenge of medical hallucination in vision language models. By enforcing rigorous agreement across heterogeneous evidence channels, specifically radiomic statistics, explainability activations, and vocabulary grounded semantic concepts, our architecture transitions from unconstrained generative inference to an auditable and responsible decision process. Quantitative evaluations on chest X ray datasets substantiate that this contextual grounding improves discriminative performance to an AUC of $0.925$ while maintaining high levels of safety compliance, even under the complexity of multi tool augmentation. Most importantly, the framework consistently yields safety aware responses and calibrated uncertainty without compromising clinically appropriate interpretations. These results confirm that richer contextual verification serves as a critical mechanism for simultaneously advancing performance and trustworthiness in medical artificial intelligence.

{
    \small
    \bibliographystyle{ieeenat_fullname}
    \bibliography{main}
}

\end{document}